\documentclass[11pt]{amsart}

\usepackage{eulervm}

\usepackage[usenames,dvipsnames,svgnames,table]{xcolor}
\usepackage{amssymb,amsfonts}
\usepackage{amsmath,amsthm}
\usepackage{mathtools}
\usepackage{pgf, tikz}
\usetikzlibrary{arrows.meta, positioning, shapes.geometric, calc, decorations.pathreplacing, fit, backgrounds, patterns}
\usepackage{enumitem}
\usepackage{graphicx}
\usepackage{booktabs}
\usepackage{tabularx}
\usepackage{dsfont}
\usepackage{multirow}
\usepackage{algorithm}
\usepackage{algorithmic}
\usepackage{caption}
\usepackage{subcaption}
\usepackage[pdfdisplaydoctitle,colorlinks,urlcolor=blue,linkcolor=blue,citecolor=blue]{hyperref}
\usepackage{listings}
\usepackage{float}
\usepackage{natbib}
\usepackage{epigraph}

\definecolor{codeblue}{RGB}{0,105,148}
\definecolor{codegray}{RGB}{100,100,100}
\definecolor{codegreen}{RGB}{0,120,60}
\definecolor{codepurple}{RGB}{120,50,150}
\definecolor{backcolor}{RGB}{248,248,248}
\definecolor{accentblue}{RGB}{30,90,150}
\definecolor{darkblue}{RGB}{20,60,110}
\definecolor{layer1}{RGB}{226,240,226}
\definecolor{layer2}{RGB}{226,232,248}
\definecolor{layer3}{RGB}{248,238,220}
\definecolor{layer4}{RGB}{244,226,226}
\definecolor{reconcile}{RGB}{215,225,245}

\lstdefinelanguage{yaml}{
  keywords={apiVersion, kind, metadata, spec, sources, access, freshness, routing, operator, trust, reliability, name, namespace, labels, type, config, refresh, ingestion, roles, agentPermissions, crossDomain, defaults, overrides, priority, intentParsing, tokenBudget, knowledgeStore, intelligence, guardrails, guardrailPolicies, anomalyDetection, audit, certification},
  keywordstyle=\color{codeblue}\bfseries,
  sensitive=true,
  comment=[l]{\#},
  commentstyle=\color{codegreen}\itshape,
  stringstyle=\color{codepurple},
  morestring=[b]',
  morestring=[b]",
}

\newtheorem{theorem}{Theorem}[section]

\theoremstyle{definition}
\newtheorem{definition}[theorem]{Definition}

\newtheorem{requirement}[theorem]{Requirement}
\newtheorem{designinvariant}[theorem]{Design Invariant}
\newtheorem{designgoal}[theorem]{Design Goal}
\theoremstyle{remark}

\numberwithin{equation}{section}

\begin{document}

\title[Context Kubernetes]
{Context Kubernetes:\\
Declarative Orchestration of Enterprise Knowledge\\for Agentic AI Systems}

\author{Charafeddine Mouzouni}

\address{OPIT -- Open Institute of Technology, and Cohorte AI, Paris, France.}
\email{charafeddine{@}cohorte.co}
\dedicatory{\small\textit{OPIT -- Open Institute of Technology, and Cohorte AI, Paris, France.}\\[2pt]\texttt{charafeddine@cohorte.co}}

\date{April 2026; revised November 2026}

\begin{abstract}
Context Kubernetes is an architecture for governing how organizational knowledge reaches AI agents. It adapts the declarative manifest and reconciliation loop of container orchestration and adds a permission model in which an agent's authority is a strict subset of its user's and high-risk actions need approval over a channel the agent cannot read or write. We implement it as a 3{,}711-line Python prototype and evaluate it on a synthetic corpus of eight documents. In our own first evaluation, the connectors stamped every file with their own role labels, so the governed pipeline sent HR and finance content to a sales agent in most queries while every permission check passed. With correct labels, the governed pipeline and a plain access-control filter both deliver no out-of-role content, by construction. At equal output size, the filter reaches more of the queries the agent may answer (116 of 134, against 105) at the same pooled precision, so on this corpus intent routing adds little. In five attack scenarios, per-agent approval tiers can gate an action that the user's role allows. Access control is only as good as its labels, and leakage must be tested end to end.
\end{abstract}


\maketitle

\section{Introduction}\label{sec:introduction}

\noindent An AI agent pointed at a well-organized folder can do much of a knowledge worker's routine work with files and conversation \citep{anthropic2024agents}. Moving from one agent on one laptop to thousands of agents across an organization raises the questions every earlier computing primitive raised at scale:

\begin{itemize}[leftmargin=2em, itemsep=0.2em]
  \item \textbf{Scheduling}: Which knowledge reaches which agent, from which source, in what order?
  \item \textbf{Permissions}: What can each agent read, write, and execute---and how does this relate to what its human can do?
  \item \textbf{Health monitoring}: Is the knowledge the agent received current, complete, and correct?
  \item \textbf{State management}: How is organizational knowledge versioned, governed, and migrated?
  \item \textbf{Auditability}: Who accessed what, when, through which agent, with what outcome?
\end{itemize}

For virtual machines and containers, these questions were answered by an orchestration layer above the primitive rather than by improving the primitive itself. This paper asks what the corresponding layer looks like for organizational knowledge delivered to agents, builds a prototype of it, and tests where it helps and where it does not.

We introduce \textbf{Context Kubernetes}, a reference architecture for orchestrating enterprise knowledge in agentic AI systems. For the broader enterprise agentic platform in which context orchestration operates, see \citet{mouzouni2026enterprise}. We develop the structural analogy between container orchestration and context orchestration into concrete abstractions, a declarative manifest specification, and design invariants. The analogy is productive, but it is an analogy, not a formal isomorphism, and we say where it holds tightly and where it stretches.

The evaluation's main lesson is about labels. In our own first evaluation, a permission engine that was correct over the labels it received delivered HR and finance content to a sales agent in most queries, because the connectors assigned those labels by configuration rather than by source. Every permission check passed. Once the labels were correct, a plain ranked access-control filter did about as well as the architecture's intent routing. Governance of agent context therefore depends first on how units are labelled, and leakage has to be tested end to end, through the components that assign the labels.

\paragraph{Scope.} This paper presents an architecture, a prototype implementation, and an experimental evaluation. The prototype (Section~\ref{sec:implementation}) implements the Context Router, Permission Engine, CxRI connectors, reconciliation loop, audit log, and a FastAPI service exposing the Context API. The evaluation (Section~\ref{sec:evaluation}) comprises five correctness experiments and four value experiments on a synthetic corpus of eight documents. The prototype is a research artifact: it is not production-grade and has not been deployed at a real organization.

\paragraph{A working hypothesis.} We take as a working hypothesis that when a computational primitive reaches organizational scale, the layer governing its scheduling, permissions, health, state, and audit becomes durable infrastructure. Virtualization and containers followed this pattern; whether agent context will is open, and this paper does not test it.

\subsection{Contributions}

This paper makes six contributions:

\begin{enumerate}[leftmargin=2em]
  \item A \textbf{reference architecture} with six core abstractions, a declarative manifest specification, and a structural analogy to Kubernetes identifying where the mapping is tight, moderate, and loose (Sections~\ref{sec:abstractions}--\ref{sec:conceptmap}).

  \item A \textbf{three-tier agent permission model} built on the design invariant that agent authority is a strict subset of human authority, with out-of-band strong approval architecturally isolated from the agent's execution environment (Section~\ref{sec:permissions}).

  \item A \textbf{prototype implementation} (3{,}711 lines of Python, 92 tests) comprising Context Router, Permission Engine, CxRI connectors, reconciliation loop, audit log, and HTTP API (Section~\ref{sec:implementation}).

  \item \textbf{Five correctness experiments} on the prototype's components: intent routing accuracy, seven targeted permission-engine tests, freshness-state detection latency, latency overhead, and enforcement of approval isolation (Section~\ref{sec:correctness}).

  \item \textbf{An end-to-end leakage study.} A permission engine that is correct over the labels it receives leaked HR and finance content into most queries when connectors assigned the labels. With correct labels, a ranked access-control filter is comparable to intent routing on our corpus, which places the value of governance in labelling rather than routing. We also report content-level freshness scenarios and attack scenarios that show what per-agent approval tiers can gate (Section~\ref{sec:value}).

  \item A \textbf{platform approval survey} finding that none of four major enterprise agentic platforms, as publicly documented, enforces out-of-band approval isolation (Section~\ref{sec:permissions}, Table~\ref{tab:approval_survey}).
\end{enumerate}

\subsection{Paper Organization}

Section~\ref{sec:background} establishes the container orchestration precedent and the enterprise agentic landscape. Section~\ref{sec:architecture} presents the architecture. Section~\ref{sec:declarative} introduces declarative context management. Section~\ref{sec:conceptmap} develops the structural analogy. Section~\ref{sec:implementation} describes the prototype. Section~\ref{sec:evaluation} presents the experimental evaluation. Section~\ref{sec:related} reviews related work. Section~\ref{sec:discussion} discusses limitations and open problems. Section~\ref{sec:conclusion} concludes.

\section{Background and Motivation}\label{sec:background}

\subsection{The Container Orchestration Precedent}\label{sec:containers}

Docker (2013) standardized the container as a portable unit of compute. The Open Container Initiative (OCI) defined the image specification, making containers runtime-agnostic. But deploying containers at organizational scale---scheduling across heterogeneous machines, service discovery, load balancing, rolling updates, secrets management, and access control---created problems that no individual container could solve.

Kubernetes \citep{burns2016borg, verma2015borg} addressed them with two design patterns inherited from Borg:

\paragraph{Declarative desired-state management.} Operators specify \emph{what} they want (``3 replicas of this service, 512MB memory, accessible on port 443''), not \emph{how} to achieve it. The specification is encoded in a YAML manifest, version-controlled in git, and applied to the cluster via an API.

\paragraph{The reconciliation loop.} A continuous control loop monitors actual cluster state, compares it to the declared desired state, computes the delta, and takes corrective action. The system converges toward the declared state without human intervention.

These two patterns---desired-state declaration and continuous reconciliation---are general. Nothing in them is specific to containers. This paper applies them to organizational knowledge.

\subsection{The Enterprise Agentic Landscape}\label{sec:agentic_landscape}

By early 2026, every major enterprise platform vendor has shipped an agentic AI product. Table~\ref{tab:vendors} summarizes the landscape.

\begin{table}[H]
\centering
\caption{Enterprise agentic platforms as of Q1 2026}
\label{tab:vendors}
\small
\begin{tabularx}{\textwidth}{l X l}
\toprule
\textbf{Platform} & \textbf{Architecture} & \textbf{Lock-in} \\
\midrule
Microsoft Copilot Studio & Semantic Kernel, Power Platform governance, M365 integration & High \\
Salesforce Agentforce & Atlas engine, ``System 2'' reasoning, cooperative swarms & High \\
Google Vertex AI + Agentspace & Cloud-native, Gemini models, A2A protocol contributor & Med--High \\
AWS Bedrock Agents & 100+ FMs, ReAct loop, 6 guardrail types & Medium \\
NVIDIA AI Agent Platform & NIM microservices, NeMo Guardrails, infrastructure layer & Medium \\
\midrule
LangGraph (LangChain) & Graph-based state machines, human-in-the-loop & Low \\
CrewAI & Role-based agents with tasks and crews & Low \\
AutoGen (Microsoft) & Conversational multi-agent, event-driven & Low \\
OpenAI Agents SDK & Lightweight multi-agent with handoff primitives & Low \\
\bottomrule
\end{tabularx}
\end{table}

Every platform implements some version of a four-tier architecture: (1)~reasoning engine, (2)~tool/action layer, (3)~memory/context layer, and (4)~governance layer. The protocol stack is converging under the Agentic AI Foundation at the Linux Foundation: MCP \citep{anthropic2025mcp} for agent-to-tool communication, A2A \citep{google2026a2a} for agent-to-agent communication.

Across these products, the governance layer is the least standardized and the most tied to one vendor \citep{mouzouni2025accountable}: Copilot Studio's governance does not reach a LangGraph agent, Salesforce's Einstein Trust Layer does not audit a Bedrock agent, and open-source frameworks provide orchestration without governance. A vendor-neutral governance layer for agent context is the gap this paper addresses.

\subsection{Requirements for Context Orchestration}\label{sec:requirements}

From an analysis of the vendor landscape, we identify seven requirements for a context orchestration system. We ground these in the NIST AI Risk Management Framework (AI RMF) \citep{nist2023airm} categories---Govern, Map, Measure, Manage---to ensure they are independently motivated rather than self-serving.

\begin{requirement}[Vendor Neutrality \textnormal{(NIST: Govern 1.2 --- organizational context)}]
\label{req:neutral}
The system must operate with any LLM, any agent framework, any cloud provider, and any data source.
\end{requirement}

\begin{requirement}[Declarative Management \textnormal{(NIST: Govern 1.5 --- documentation and processes)}]
\label{req:declarative}
The knowledge architecture must be expressible as a machine-readable, version-controlled specification.
\end{requirement}

\begin{requirement}[Agent Permission Separation \textnormal{(NIST: Govern 1.7 --- delineation of authority)}]
\label{req:perms}
Agent permissions must be formally separated from user permissions. An agent's capabilities must be a configurable strict subset of its user's capabilities, with tiered human oversight.
\end{requirement}

\begin{requirement}[Context Freshness \textnormal{(NIST: Measure 2.6 --- data quality and relevance)}]
\label{req:freshness}
The system must continuously monitor the currency of all knowledge and take corrective action when knowledge exceeds its configured time-to-live.
\end{requirement}

\begin{requirement}[Intent-Based Access \textnormal{(NIST: Map 1.5 --- information management)}]
\label{req:intent}
Agents must request knowledge by intent, not by location. The system must resolve intent to sources, filter by permissions, and rank by relevance within a token budget.
\end{requirement}

\begin{requirement}[Complete Auditability \textnormal{(NIST: Manage 4.2 --- documentation of incidents and actions)}]
\label{req:audit}
Every knowledge access, every action, and every approval must be immutably logged with attribution, timestamp, and outcome.
\end{requirement}

\begin{requirement}[Organizational Intelligence \textnormal{(NIST: Manage 4.1 --- continuous improvement)}]
\label{req:intelligence}
The system must accumulate cross-organizational knowledge from aggregate agent activity, without exposing individual data across permission boundaries.
\end{requirement}

\section{Context Kubernetes: Architecture}\label{sec:architecture}

\subsection{Design Principles}\label{sec:principles}

Context Kubernetes is governed by seven design principles, each traceable to a requirement:

\begin{enumerate}[leftmargin=2em]
  \item \textbf{Declarative over imperative} (Req.~\ref{req:declarative}). Organizations declare the knowledge architecture they want. The system reconciles reality to match.
  \item \textbf{Intent-based access over location-based} (Req.~\ref{req:intent}). Agents request knowledge by semantic intent. The system resolves intent to sources, permissions, and rankings.
  \item \textbf{Governance as foundation, not afterthought} (Req.~\ref{req:perms}, \ref{req:freshness}, \ref{req:audit}). Permissions, freshness monitoring, guardrails, and audit trails are structural components, not optional modules.
  \item \textbf{Agent authority as strict subset of human authority} (Req.~\ref{req:perms}). For actions, the Permission Engine registers an agent profile only if it is a strict subset of its user's role; the prototype does not yet apply the profile to reads (Section~\ref{sec:limitations}).
  \item \textbf{Intelligence in the orchestration layer} (Req.~\ref{req:intent}, \ref{req:intelligence}). The routing layer is designed for LLM-assisted semantic parsing. Context Operators accumulate organizational intelligence. Unlike Kubernetes, the orchestration layer itself reasons---a property we examine critically in Section~\ref{sec:llm_tension}.
  \item \textbf{Vendor neutrality} (Req.~\ref{req:neutral}). Every component is defined by its interface, not its implementation.
  \item \textbf{Privacy by design.} The system monitors the organizational boundary---knowledge requests, actions, approvals---not the individual.
\end{enumerate}

\subsection{Core Abstractions}\label{sec:abstractions}

We define six abstractions that constitute the foundational vocabulary of context orchestration.

\begin{definition}[Context Unit]
\label{def:cu}
A \emph{context unit} is the smallest addressable element of organizational knowledge. Formally, $u = (c, \tau, m, v, e, \pi)$ where $c$ is the content, $\tau \in \{\texttt{unstructured}, \texttt{structured}, \texttt{hybrid}\}$ is the type, $m$ is a metadata set (author, timestamp, domain, sensitivity, entities, source), $v \in \mathcal{V}$ is a version identifier, $e \in \mathbb{R}^d$ is an embedding vector, and $\pi \subseteq \mathcal{R}$ is the set of roles authorized to access this unit.
\end{definition}

\begin{definition}[Context Domain]
\label{def:cd}
A \emph{context domain} is an isolation boundary for organizational knowledge: $\mathcal{D} = (N, S, A, F, \rho, O, G)$ where $N$ is the domain identifier, $S$ is the set of backing sources, $A: \mathcal{R} \rightarrow 2^{\texttt{Ops}} \times \texttt{Tier}$ is the access control function, $F$ is the freshness policy, $\rho$ is the routing configuration, $O$ is the domain operator, and $G$ is the guardrail set. A query within domain $\mathcal{D}_i$ has no visibility into domain $\mathcal{D}_j$ unless explicitly brokered through the target domain's operator with the requester's permissions propagated.
\end{definition}

\begin{definition}[Context Store]
\label{def:cs}
A \emph{context store} is a backing system that persists context units: $s = (\sigma, \iota, \phi)$ where $\sigma$ is the store type (git repository, relational database, connector to external system, file system), $\iota$ is the ingestion configuration, and $\phi$ is the connection specification. Stores are accessed exclusively through the Context Runtime Interface.
\end{definition}

\begin{definition}[Context Endpoint]
\label{def:ce}
A \emph{context endpoint} $\varepsilon$ is a stable, intent-based interface for knowledge access. Given intent $q$, session scope $\omega$, and agent permission profile $\alpha$:
\[
  \varepsilon(q, \omega, \alpha) \rightarrow \{u_1, \ldots, u_k\} \subseteq \mathcal{U}
\]
subject to $\forall u_i$: $\pi(u_i) \ni \text{role}(\alpha)$, $\text{fresh}(u_i)$, and $\sum_i |u_i| \leq B$ where $B$ is the token budget. The agent never specifies \emph{where} knowledge lives---only \emph{what} it needs.
\end{definition}

\begin{definition}[Context Runtime Interface (CxRI)]
\label{def:cxri}
The CxRI is a standard adapter between the orchestration layer and context stores. Every store implements six operations:
\begin{align*}
  &\texttt{connect}(\phi) \rightarrow \texttt{Connection}
  &&\texttt{query}(\texttt{conn}, q) \rightarrow \{u_1, \ldots, u_n\} \\
  &\texttt{read}(\texttt{conn}, \texttt{path}) \rightarrow u
  &&\texttt{write}(\texttt{conn}, \texttt{path}, c) \rightarrow \texttt{Result} \\
  &\texttt{subscribe}(\texttt{conn}, \texttt{path}) \rightarrow \texttt{Stream}
  &&\texttt{health}(\texttt{conn}) \rightarrow \texttt{Status}
\end{align*}
The CxRI ensures that the orchestration layer is decoupled from any specific data source, analogous to Kubernetes' Container Runtime Interface (CRI).
\end{definition}

\begin{definition}[Context Operator]
\label{def:co}
A \emph{context operator} is a domain-specific autonomous controller: $O = (K, L, I, \Gamma)$ where $K$ is a knowledge store (vector + full-text index), $L$ is a reasoning engine (LLM with domain guardrails), $I$ is an organizational intelligence module, and $\Gamma$ is the guardrail set. Operators extend the base system with domain intelligence, inspired by Kubernetes Operators \citep{dobies2020operators} that extend the base scheduler with application-specific logic.

The intelligence module $I$ extracts cross-organizational patterns from aggregate agent activity, subject to: minimum signal threshold $\theta \geq 3$ (preventing premature pattern recognition), temporal clustering within window $\Delta t_w$, permission tagging (aggregated insights available to domain members; attributed insights require record-level authorization), and decay (insights degrade without supporting signals). The governance implications of this module are discussed in Section~\ref{sec:intelligence_governance}.
\end{definition}

\subsection{System Architecture}\label{sec:components}

\begin{figure}[H]
\centering
\begin{tikzpicture}[
  cbox/.style={draw, rounded corners=2pt, minimum width=2.0cm, minimum height=0.7cm, align=center, font=\scriptsize},
  sbox/.style={draw, rounded corners=2pt, minimum width=1.6cm, minimum height=0.7cm, align=center, font=\scriptsize},
  bar/.style={draw, rounded corners=3pt, minimum width=12.0cm, minimum height=0.5cm, align=center, font=\scriptsize\bfseries},
  layer/.style={draw, rounded corners=3pt, minimum width=12.0cm, align=center},
  lbl/.style={font=\scriptsize\bfseries, text=darkblue},
  >=Stealth
]

\node[layer, fill=layer4!55, minimum height=1.9cm] (agents) at (0, 10.85) {};
\node[lbl, anchor=north west] at ([shift={(5pt,-4pt)}]agents.north west) {AGENTS};
\node[cbox, fill=white] at (-3.75, 10.55) {Local\\Agent};
\node[cbox, fill=white] at (-1.25, 10.55) {Claude\\Code};
\node[cbox, fill=white] at (1.25, 10.55) {LangGraph\\App};
\node[cbox, fill=white] at (3.75, 10.55) {Custom\\Agent};

\node[bar, fill=accentblue!12] (api) at (0, 9.35) {Context API \quad (10 endpoints $\cdot$ HTTPS + WSS $\cdot$ JWT)};

\node[layer, fill=layer2!45, minimum height=4.6cm] (ck8s) at (0, 6.45) {};
\node[lbl, anchor=north west] at ([shift={(5pt,-4pt)}]ck8s.north west) {CONTEXT KUBERNETES};

\node[cbox, fill=white] at (-3.75, 7.55) {Context\\Registry};
\node[cbox, fill=white] at (-1.25, 7.55) {Context\\Router};
\node[cbox, fill=white] at (1.25, 7.55) {Permission\\Engine};
\node[cbox, fill=white] at (3.75, 7.55) {Freshness\\Manager};

\node[cbox, fill=white] at (-3.75, 6.45) {Trust Policy\\Engine};
\node[cbox, fill=white] at (-1.25, 6.45) {Context\\Operators};
\node[cbox, fill=white] at (1.25, 6.45) {Observability};
\node[cbox, fill=white] at (3.75, 6.45) {LLM\\Gateway};

\node[bar, fill=reconcile, font=\scriptsize\bfseries\itshape] (recon) at (0, 5.35) {Reconciliation Loop \quad $\langle$\,desired state\,$\overset{\Delta}{\longleftrightarrow}$\,actual state\,$\rangle$};

\node[bar, fill=layer3!60] (cxri) at (0, 3.55) {Context Runtime Interface (CxRI) \quad --- \quad 6 ops per connector};

\node[layer, fill=layer1!65, minimum height=1.9cm] (sources) at (0, 2.05) {};
\node[lbl, anchor=north west] at ([shift={(5pt,-4pt)}]sources.north west) {CONTEXT SOURCES};
\node[sbox, fill=white] at (-3.8, 1.75) {Git\\Repos};
\node[sbox, fill=white] at (-1.9, 1.75) {Databases};
\node[sbox, fill=white] at (0, 1.75) {Email /\\Calendar};
\node[sbox, fill=white] at (1.9, 1.75) {SaaS\\APIs};
\node[sbox, fill=white] at (3.8, 1.75) {ERP /\\Legacy};

\draw[->, thick] (agents.south) -- (api.north);
\draw[->, thick] (api.south) -- (ck8s.north);
\draw[->, thick] (ck8s.south) -- (cxri.north);
\draw[->, thick] (cxri.south) -- (sources.north);

\end{tikzpicture}
\caption{Context Kubernetes reference architecture.}
\label{fig:architecture}
\end{figure}
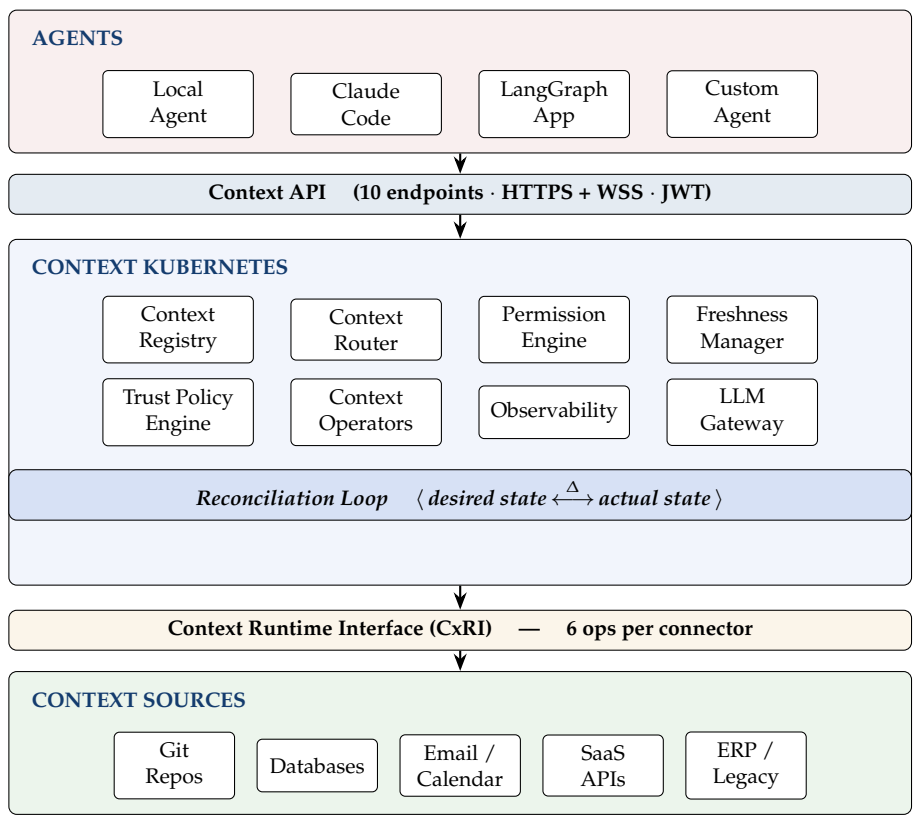

Figure~\ref{fig:architecture} presents the architecture. The system comprises seven services: Context Registry (metadata store, analogous to \texttt{etcd}), Context Router (scheduling intelligence, analogous to \texttt{kube-scheduler} + \texttt{Ingress}), Permission Engine (three-tier agent permission model, Section~\ref{sec:permissions}), Freshness Manager (continuous monitoring with four states: fresh, stale, expired, conflicted), Trust Policy Engine (declarative guardrail evaluation, anomaly detection, DLP, audit logging), Context Operators (domain-specific controllers, Def.~\ref{def:co}), and LLM Gateway (prompt inspection and cost metering for enterprise deployments).

\subsection{The Agent Permission Model}\label{sec:permissions}

Existing RBAC models \citep{sandhu1996rbac} manage human identities. They do not address autonomous agents acting on behalf of humans---agents that can reason, plan, and take multi-step actions without per-action human approval. We propose two design invariants:

\begin{designinvariant}[Agent Permission Subset]
\label{inv:subset}
For every user $u$ with permission set $P_u$ and agent $a_u$ acting on behalf of $u$:
\[
  P_{a_u} \subset P_u
\]
The inclusion is \emph{strict}: for every user, there exists at least one operation that the user can perform but the agent cannot. For every shared operation $o \in P_{a_u}$, the required approval tier satisfies $T(o, a_u) \geq T(o, u)$.
\end{designinvariant}

This invariant is enforced by construction for actions: the Permission Engine computes $P_{a_u}$ as a configured projection of $P_u$. No API exists for an agent to modify its own permission profile. We note that this invariant is a design commitment, not a formally verified property; a complete implementation would require verification that no code path allows escalation, which is a standard security engineering task but is not addressed in this paper.

The three-tier approval model (Table~\ref{tab:tiers}) enforces graduated human oversight:

\begin{table}[H]
\centering
\small
\caption{Three-tier agent approval model}
\label{tab:tiers}
\begin{tabularx}{\textwidth}{l l X l}
\toprule
\textbf{Tier} & \textbf{Agent Role} & \textbf{Approval Mechanism} & \textbf{Examples} \\
\midrule
T1: Autonomous & Acts freely & None & Read context, draft docs \\
T2: Soft approval & Proposes & User confirms in agent UI & Send internal message \\
T3: Strong approval & Surfaces task & Out-of-band 2FA / biometric & Sign contract, financials \\
Excluded & Cannot request & N/A (manual only) & Terminate employee \\
\bottomrule
\end{tabularx}
\end{table}

\begin{designinvariant}[Strong Approval Isolation]
\label{inv:isolation}
For any Tier~3 action, the approval channel $C$ satisfies: (1)~$C$ is external to the agent's execution environment, (2)~$C$ is neither readable nor writable by the agent, and (3)~$C$ requires a separate authentication factor.
\end{designinvariant}

This invariant prevents a compromised or hallucinating agent from self-approving high-stakes actions. To assess whether this gap is real, we surveyed the published security architectures of four major enterprise agentic platforms against the three conditions of Design Invariant~\ref{inv:isolation}. Table~\ref{tab:approval_survey} summarizes the findings.

\begin{table}[H]
\centering
\caption{Approval isolation survey: four enterprise agentic platforms assessed against Design Invariant~\ref{inv:isolation}. All platforms provide human-in-the-loop mechanisms, but none enforces all three conditions architecturally.}
\label{tab:approval_survey}
\small
\begin{tabularx}{\textwidth}{l X c c c}
\toprule
\textbf{Platform} & \textbf{HITL mechanism} & \rotatebox{70}{\textbf{C1: External}} & \rotatebox{70}{\textbf{C2: Not R/W}} & \rotatebox{70}{\textbf{C3: Sep.\ factor}} \\
\midrule
MS Copilot Studio & Request for Information via Outlook email; AI Approvals & Partial$^a$ & No$^b$ & No \\
\addlinespace
SF Agentforce & Guardrails (deterministic filters + LLM instructions); escalation patterns & No$^c$ & No & No \\
\addlinespace
AWS Bedrock & User Confirmation via \texttt{InvokeAgent} API; AgentCore Cedar policy & No$^d$ & No$^e$ & No \\
\addlinespace
Google Vertex AI & \texttt{require\_confirmation} in ADK; \texttt{before\_tool\_callback} & No$^f$ & No$^g$ & No \\
\bottomrule
\end{tabularx}
\vspace{0.3em}

\footnotesize
$^a$Email delivery is external, but the response flows back into the agent flow as readable parameters.
$^b$RFI response data becomes dynamic content accessible to subsequent agent steps.
$^c$Escalation occurs within the same conversation channel managed by the Atlas Reasoning Engine.
$^d$AgentCore Gateway + Cedar policies operate \emph{outside} agent code (strongest policy isolation surveyed), but the User Confirmation mechanism is in-band via the same API session.
$^e$Confirmation state is returned via \texttt{sessionState} in the same \texttt{InvokeAgent} request.
$^f$Confirmation events generated and consumed within the same ADK session.
$^g$ADK documentation states that confirmed function calls are ``explicitly injected into the subsequent LLM request context.''
\end{table}

\paragraph{Key finding.} All four platforms provide human-in-the-loop mechanisms, but in every case the approval channel is \emph{in-band} with the agent's execution or conversation context. No platform enforces \emph{channel separation} (approval delivery via a channel the agent cannot read or write). No platform requires \emph{separate-factor authentication} for the approval act itself. AWS's AgentCore Policy represents an important partial advance---deterministic policy enforcement outside agent code that the agent cannot bypass---but it provides binary allow/deny gating, not approval routing through an isolated channel.

The gap between ``external policy enforcement'' and ``external approval channel with separate-factor authentication'' is precisely the contribution of Design Invariant~\ref{inv:isolation}.

\subsection{The Reconciliation Loop}\label{sec:reconciliation}

The reconciliation loop (Algorithm~\ref{alg:reconcile}) continuously compares declared state $D$ (from the manifest) against observed state $A$ (from the registry and sources), computes drift $\Delta$, and executes corrective actions.

\begin{algorithm}[H]
\caption{Context Reconciliation Loop}
\label{alg:reconcile}
\begin{algorithmic}[1]
\LOOP
  \STATE $D \leftarrow \textsc{ReadManifest}()$
  \STATE $A \leftarrow \textsc{ObserveState}(\text{registry}, \text{sources}, \text{operators})$
  \STATE $\Delta \leftarrow \{(d, a) \mid d \in D, a \in A, d \neq a\}$
  \FOR{each $\delta = (\text{type}, \text{target}) \in \Delta$}
    \IF{$\delta.\text{type} = \texttt{source\_disconnected}$}
      \STATE Retry $\to$ mark dependents stale $\to$ alert
    \ELSIF{$\delta.\text{type} = \texttt{context\_stale}$}
      \STATE Apply freshness policy: re-sync $|$ flag $|$ archive
    \ELSIF{$\delta.\text{type} = \texttt{operator\_unhealthy}$}
      \STATE Restart operator $\to$ alert
    \ELSIF{$\delta.\text{type} = \texttt{anomaly}$}
      \STATE Evaluate policy $\to$ alert $|$ throttle $|$ suspend
    \ELSIF{$\delta.\text{type} = \texttt{reliability\_drift}$}
      \STATE Trigger recertification \citep{mouzouni2026trustgate}
    \ELSIF{$\delta.\text{type} = \texttt{permission\_change}$}
      \STATE Propagate to active sessions
    \ENDIF
  \ENDFOR
  \STATE \textsc{UpdateRegistry}($A$)
  \STATE \textbf{wait} $\Delta t_r$
\ENDLOOP
\end{algorithmic}
\end{algorithm}

We state two intended design goals for the reconciliation loop. These are not formally proven; they are design commitments that an implementation must satisfy and that should be verified through testing and model checking. Section~\ref{sec:limitations} states what has been checked so far.

\begin{designgoal}[Safety]
\label{goal:safety}
The reconciliation loop should preserve: (1)~\textbf{Permission safety}: no context unit is delivered to an agent whose permission profile does not include the unit's access scope; if the Permission Engine is unavailable, all requests are denied (fail-closed). (2)~\textbf{Freshness safety}: no context unit past its \texttt{expired} threshold is served; stale units are served with explicit staleness metadata.
\end{designgoal}

\begin{designgoal}[Liveness]
\label{goal:liveness}
Under the assumption that at least one CxRI connector per domain remains reachable: (1)~every stale context unit is re-synced or flagged within $2 \cdot \Delta t_r$, and (2)~every disconnected source is detected within $\Delta t_r$.
\end{designgoal}

\section{Declarative Context Management}\label{sec:declarative}

\subsection{Knowledge Architecture as Code}\label{sec:kac}

Infrastructure-as-code tools made compute configuration declarative and reviewable, and Kubernetes made that the norm for running workloads. Context Kubernetes proposes the same for knowledge, \textbf{Knowledge Architecture as Code}: the organizational knowledge landscape---sources, permissions, freshness policies, routing rules, trust policies, and operator configurations---expressed as a declarative manifest, version-controlled in git, reviewed before application, and continuously reconciled.

\subsection{The Context Architecture Manifest}\label{sec:manifest}

The manifest specification comprises seven sections. Listing~\ref{lst:manifest} shows a representative domain declaration.

\begin{lstlisting}[caption={Context Domain Manifest for a Sales domain},label={lst:manifest}]
apiVersion: context/v1
kind: ContextDomain
metadata:
  name: sales
  namespace: acme-corp
  labels: {sensitivity: confidential, owner: head-of-sales}

spec:
  sources:
    - name: client-context
      type: git-repo
      config: {repo: git@ctx.internal:sales/clients.git}
      refresh: realtime
      ingestion: {chunking: semantic, chunkSize: 500,
                  embedding: text-embedding-3-small}
    - name: pipeline
      type: connector
      config: {system: salesforce, scope: opportunities,
               credentials: vault://sf/key}
      refresh: 1h
    - name: communications
      type: connector
      config: {system: gmail, filter: "label:client-comms"}
      refresh: 15m
      ingestion: {chunking: per-thread, ttl: 180d}

  access:
    roles:
      - role: sales-rep
        read: ["clients/${assigned}/*"]
        write: ["clients/${assigned}/*"]
      - role: sales-manager
        read: ["*"]
        write: ["*"]
    agentPermissions:
      read: autonomous
      write:
        default: soft-approval
        paths:
          "*/contracts/*": strong-approval
          "pipeline/*": autonomous
      execute:
        send-internal-msg: soft-approval
        send-external-email: strong-approval
        commit-to-pricing: excluded
    crossDomain:
      - {domain: operations, mode: brokered}
      - {domain: finance, mode: brokered}
      - {domain: hr, mode: denied}

  freshness:
    defaults: {maxAge: 24h, staleAction: flag}
    overrides:
      - {path: "*/communications/*", maxAge: 4h,
         staleAction: re-sync}
      - {path: "pipeline/*", maxAge: 1h,
         staleAction: re-sync}

  routing:
    intentParsing: llm-assisted
    tokenBudget: 8000
    priority:
      - {signal: semantic_relevance, weight: 0.40}
      - {signal: recency,           weight: 0.30}
      - {signal: authority,          weight: 0.20}
      - {signal: user_relevance,     weight: 0.10}

  operator:
    type: master-agent
    template: sales-v2
    intelligence:
      patternEngine: {minSignals: 3, window: 30d}
    guardrails:
      - "CANNOT commit to pricing without approval"
      - "CANNOT share client A data with client B"

  trust:
    policies:
      - name: no-unreviewed-external-email
        trigger: action.send_email
        condition: recipient.domain != company.domain
        action: require_approval(tier: strong)
    anomalyDetection:
      baseline: per-user-per-role
      threshold: 3x
      response: alert-admin
    audit: {level: full, retention: 7y}

  reliability:
    minLevel: 0.90
    method: trustgate
    schedule: {deploy: true, model-change: true,
               monthly: true}
\end{lstlisting}

\subsection{Context Migration as Rolling Update}\label{sec:migration}

Knowledge migration follows the Kubernetes rolling-update pattern: (1)~\textbf{Connect}: attach CxRI connectors, all data stays in place; (2)~\textbf{Duplicate \& Shift}: extract narrative data into git-versioned context files, both sources in parallel with incremental sync; (3)~\textbf{Consolidate}: sunset redundant tools, keep essential ones as connectors; (4)~\textbf{Steady state}: organization operates on Context Kubernetes, no agent code changes during migration.

\section{The Structural Analogy}\label{sec:conceptmap}

Table~\ref{tab:map} presents the complete mapping between Kubernetes primitives and their context orchestration counterparts. We use the term \emph{structural analogy} rather than \emph{isomorphism}: many mappings are tight functional equivalences (CRI~$\leftrightarrow$~CxRI, RBAC~$\leftrightarrow$~Agent Permission Profile), but others are looser design inspirations (Service Mesh~$\leftrightarrow$~Trust Layer, HPA~$\leftrightarrow$~Context Caching). We mark the tightness of each mapping explicitly.

\begin{table}[H]
\centering
\caption{Structural analogy: Kubernetes $\leftrightarrow$ Context Kubernetes. Tightness: \textbf{T}ight (functional equivalence), \textbf{M}oderate (shared pattern, different mechanics), \textbf{L}oose (design inspiration).}
\label{tab:map}
\small
\begin{tabularx}{\textwidth}{l l X c}
\toprule
\textbf{Kubernetes} & \textbf{Context K8s} & \textbf{Shared function} & \\
\midrule
Pod & Context Unit & Smallest schedulable/addressable unit & T \\
Namespace & Context Domain & Isolation boundary with scoped access & T \\
Deployment & Context Architecture & Declared desired state & T \\
RBAC & Agent Perm.\ Profile & Role $\to$ capabilities (extended with tiers) & T \\
CRI & CxRI & Standard runtime adapter & T \\
etcd & Context Registry & Authoritative metadata store & T \\
kubectl / API & Context API & Programmatic control interface & T \\
\midrule
Ingress & Context Router & Routes external requests to internal resources & M \\
Persistent Volume & Context Store & Durable storage abstracted by interface & M \\
Liveness Probe & Freshness Manager & Health monitoring with corrective action & M \\
Rolling Update & Context Migration & Gradual replacement preserving availability & M \\
Operator & Context Operator & Domain-specific controller extending base system & M \\
Service & Context Endpoint & Stable abstract interface & M \\
\midrule
ConfigMap/Secret & Context Classification & Metadata governing handling & L \\
HPA & Context Caching & Demand-based scaling & L \\
Service Mesh & Trust Layer & Policy enforcement and observability & L \\
Helm Chart & Architecture Template & Packaged configurations & L \\
\bottomrule
\end{tabularx}
\end{table}

\subsection{Where Context Orchestration Extends Beyond Containers}\label{sec:extensions}

Four properties make context orchestration harder than container orchestration:

\textbf{1. Heterogeneity.} Containers are standardized (OCI image spec). Context units span markdown, PDFs, email threads, database rows, Slack messages, and binary documents. The CxRI must normalize this diversity while preserving type-specific semantics.

\textbf{2. Semantics.} Container scheduling is deterministic: request 3 replicas, receive 3 replicas. Context routing requires \emph{judgment}: ``Get me the Henderson context'' demands disambiguation, role-aware scoping, and relevance ranking. This is why the Context Router embeds LLM-assisted intelligence---the orchestration layer \emph{itself} reasons. This is also the architecture's deepest tension, discussed in Section~\ref{sec:llm_tension}.

\textbf{3. Sensitivity.} Containers are access-neutral. Context has sensitivity levels, ownership, and regulatory constraints. The trust layer has no Kubernetes equivalent.

\textbf{4. Learning.} Containers are stateless by design. Context Operators accumulate organizational intelligence. The system accumulates state over time---a property that has no container analog and raises novel governance questions (Section~\ref{sec:intelligence_governance}).

Table~\ref{tab:properties} maps each property to the component that addresses it and the open problem that remains.

\begin{table}[H]
\centering
\caption{Four distinguishing properties: architectural response and open problems}
\label{tab:properties}
\small
\begin{tabular}{@{} l p{4.2cm} p{7.8cm} @{}}
\toprule
\textbf{Property} & \textbf{Component \& mechanism} & \textbf{Open problem} \\
\midrule
Heterogeneity & CxRI + ingestion pipelines normalize all sources into uniform context units & No standard for context unit serialization (cf.\ OCI image spec); cross-type semantic alignment unsolved \\
\addlinespace
Semantics & Context Router with LLM-assisted intent parsing + multi-signal ranking & Probabilistic routing can violate governance guarantees; fallback degrades on ambiguous queries (\S\ref{sec:llm_tension}) \\
\addlinespace
Sensitivity & Permission Engine + Trust Policy Engine: strict-subset invariant, 3-tier approval, DLP & Formal verification of non-escalation across chained workflows; side channels via ranking/truncation \\
\addlinespace
Learning & Context Operators: pattern extraction ($\theta \geq 3$), decay, domain-owner review & Emergent sensitivity of aggregates; legal discoverability; feedback loops (\S\ref{sec:intelligence_governance}) \\
\bottomrule
\end{tabular}
\end{table}

\section{Implementation}\label{sec:implementation}

\enlargethispage{2pt}We implement Context Kubernetes as an open-source Python prototype.\footnote{Source code: \url{https://github.com/Cohorte-ai/context-kubernetes}, tag \texttt{arxiv-v4}.} The implementation comprises seven components totaling 3{,}711 lines (5{,}471 including the 92 automated tests):

\begin{itemize}[leftmargin=2em, itemsep=0.1em]
  \item \textbf{Context Router}: rule-based intent classification with 7-domain keyword taxonomy (25+ keywords per domain), multi-signal ranking (recency, semantic relevance, authority, user relevance), token budget enforcement, intent caching, conversation-aware co-reference resolution.
  \item \textbf{Permission Engine}: three-tier approval model, strict-subset invariant enforcement at registration time, OTP-based Tier~3 with out-of-band channel simulation, session management with kill switches (single/user/global).
  \item \textbf{CxRI connectors}: Git repository (6 operations, file-to-ContextUnit conversion with git metadata), PostgreSQL (table search, schema discovery).
  \item \textbf{Reconciliation loop}: freshness state machine (fresh/stale/expired), source health monitoring, consecutive-failure thresholds, event callbacks, configurable re-sync policies.
  \item \textbf{Audit log}: append-only event recording (in-memory and JSONL file backends), queryable by session, user, event type.
  \item \textbf{Manifest parser}: loads YAML domain manifests into typed Python objects.
  \item \textbf{FastAPI service}: 10 HTTP endpoints implementing the Context API (sessions, context requests, action submission, approval resolution, audit queries, health checks).
\end{itemize}

The intent classifier can call a small LLM (gpt-4o-mini) and falls back to keyword rules. All experiments in this paper use the rule-based classifier, so they are reproducible without API access; Section~\ref{sec:llm_tension} reports what the LLM classifier did on our queries.

Seed data for the experiments is synthetic: a 10-person consulting firm with three clients, five context domains (clients, sales, delivery, HR, finance), and eight context files.

\section{Evaluation}\label{sec:evaluation}

We evaluate Context Kubernetes through nine experiments in two categories: five \emph{correctness experiments} that check that individual components behave as specified, and four \emph{value experiments} (V1, V1b, V2, V3) that examine end-to-end behaviour. All experiments run against the prototype on the eight-document synthetic corpus; with a corpus this small, they show mechanisms and failure modes, not performance at scale.

\subsection{Correctness Experiments}\label{sec:correctness}

\textbf{C1: Routing quality.} 47 queries across 5 domains with ground-truth relevance labels. The rule-based router achieves 70.2\% domain classification accuracy and 0.626 MRR (Mean Reciprocal Rank); on the 200-query set used in V1b, accuracy is 63\%. Average routing latency is 0.2--0.4\,s, depending on machine load.

\textbf{C2: Permission correctness.} 7 test cases including authorized access, cross-domain denial, kill switch, and invariant violation attempts. Result: \textbf{zero unauthorized deliveries, zero false positives, zero invariant violations} across the seven cases. C2 tests the Permission Engine directly, on units whose labels are given; it does not exercise how connectors assign those labels. V1 shows why that matters.

\textbf{C3: Freshness behavior.} 9 scenarios testing the reconciliation loop's source-level freshness states: stale detection, disconnect detection, recovery, re-sync triggering, and multi-source independence. Stale detection latency: 0.65\,ms. Disconnect detection: 0.02\,ms. Reconciliation cycle for 20 sources: 23\,ms average. All state transitions (fresh $\to$ stale $\to$ expired) verified.

\textbf{C4: Latency overhead.} Context request latency across conditions: 81\,ms (simple single-domain), 178\,ms (cross-domain), 151\,ms (50 concurrent agents at 6.6 queries/second). Direct file read baseline: 0.02\,ms. The overhead is dominated by CxRI connector I/O and intent classification, not permission checking.

\textbf{C5: Approval isolation.} 8 tests verifying Design Invariant~\ref{inv:isolation}: OTP never appears in any API response accessible to the agent, 100 wrong OTP attempts all rejected, Tier~3 cannot be resolved through the Tier~2 path, replay attacks fail, expired OTPs rejected, kill switch blocks all subsequent actions. Result: \textbf{8/8 pass.} Within the prototype, where the out-of-band channel is simulated, the agent can neither read nor bypass the approval step.

\subsection{Value Experiments}\label{sec:value}

The value experiments examine end-to-end behaviour: what reaches an agent with and without governance (V1, V1b), what goes wrong when content is out of date (V2), and which actions each permission model gates (V3).

\textbf{V1: Context delivery and connector labels.} The same 47 queries, issued by an agent acting for a sales representative, are processed through two pipelines: (a)~an ungoverned pipeline (keyword retrieval over the whole repository, top-20, no permissions, no freshness) and (b)~Context Kubernetes (intent routing, per-domain connectors, permission filtering, multi-signal ranking). A delivered unit is a \emph{leak} if it comes from a domain the sales role may not read (HR or finance). The keyword retriever is crude: it counts a file as a match if any query word, function words included, occurs in its text or path, so it returns about seven of the eight files per query. Ungoverned retrieval therefore reaches HR or finance content in almost every query (46 of 47, and 196 of 200 on the larger set); that figure describes this harness, not keyword retrieval in general.

Units carry the roles of the connector that fetches them. A connector configured without a path scope therefore serves the whole repository under its own labels: the ``sales'' connector delivers \texttt{hr/policies.md} labelled as readable by sales, and the Permission Engine, correct over the labels it is given, lets it through. This is the configuration we first evaluated. Table~\ref{tab:gov_vs_ungov} compares it with two correct ones: each connector scoped to its own folder, and the same with client profiles added as a second, correctly labelled source of the sales domain (the intent classifier routes client queries to sales).

We count empty responses separately and report two precisions: \emph{per-query} precision, the mean over non-empty responses of the share of delivered units that are relevant, and \emph{pooled} precision, all relevant units delivered over all units delivered. Twelve of the 47 queries have relevant files only in HR or finance, which a sales agent must not see; the remaining 35 are the queries a correct pipeline can answer, and ``permitted reached'' counts those for which a relevant, permitted unit was delivered.

\begin{table}[H]
\centering
\caption{Context delivery to a sales agent, 47 queries, synthetic eight-document corpus. ``Unscoped'' is the configuration evaluated in the April 2026 versions of this paper.}
\label{tab:gov_vs_ungov}
\small
\begin{tabular}{@{} l r r r r @{}}
\toprule
 & & \multicolumn{3}{c}{\textbf{Context Kubernetes}} \\
\cmidrule(l){3-5}
\textbf{Metric} & \textbf{Ungoverned} & \textbf{Unscoped} & \textbf{Scoped} & \textbf{Scoped + clients} \\
\midrule
Units delivered per query & 7.1 & 6.3 & 0.8 & 3.1 \\
Leaked units & 87 & 76 & 0 & 0 \\
Queries with at least one leak & 46 & 38 & 0 & 0 \\
Empty responses & 0 & 8 & 13 & 9 \\
Precision, per query (non-empty) & 18.6\% & 18.6\% & 60.3\% & 35.6\% \\
Precision, pooled & 18.5\% & 18.5\% & 58.3\% & 25.9\% \\
Permitted queries reached (of 35) & 34 & 32 & 21 & 28 \\
\bottomrule
\end{tabular}
\end{table}

Three observations. First, unscoped, the governed pipeline leaks into 38 of 47 queries, and whenever it returns anything it is no more precise than ungoverned retrieval; its 76 leaked units are HR policy (37) and finance budgets (39). Second, with scoped connectors no out-of-role unit is delivered. This follows from the labels: the permission check and the leak definition test the same table of domains and roles, so once each connector labels units by folder, zero leaks holds by construction. It shows that the configuration is correct, not that routing prevents leaks. Third, scoping exposes a recall cost that leakage had hidden: a misrouted query now returns nothing useful. Scoping alone reaches 21 of the 35 answerable queries; adding client profiles as a sales source reaches 28.

\textbf{V1b: Matched baselines, 200 queries.} The 200-query set adds 153 templated paraphrases to the 47 queries, over the same eight files. Unscoped, both governed pipelines leak here too (295 and 307 units, in 155 of the 200 queries); Table~\ref{tab:fourway} uses correctly labelled connectors and adds two kinds of baseline. An \emph{ACL filter} retrieves from the whole repository and drops units outside the role; its zero leaks also hold by construction, since it drops units with the leak predicate itself. Because the filter returns about 5.4 of the 8 documents per query and the governed pipeline 2.8, and a pipeline that returns fewer units tends to score higher on precision, we also truncate the filter to its top $k$ units in the connector's ranking. An \emph{RBAC} pipeline uses the same router and connectors as Context Kubernetes without its cross-domain rule; the two deliver context almost identically, because the three-tier model governs actions, not reads (V3 tests it).

\begin{table}[H]
\centering
\caption{Matched baselines, 200 queries, correctly labelled connectors. RBAC and Context Kubernetes use scoped connectors plus the client source. Of the 200 queries, 134 have a relevant file a sales agent may read. Ties in the filter's ranking are broken by file path.}
\label{tab:fourway}
\small
\setlength{\tabcolsep}{4pt}
\begin{tabular}{@{} l r r r r r r r @{}}
\toprule
 & & \multicolumn{4}{c}{\textbf{ACL filter}} & & \\
\cmidrule(lr){3-6}
 & \textbf{Ungov.} & \textbf{all} & \textbf{top-4} & \textbf{top-3} & \textbf{top-2} & \textbf{RBAC} & \textbf{CK8s} \\
\midrule
Units delivered per query & 7.3 & 5.4 & 3.8 & 2.9 & 2.0 & 2.8 & 2.8 \\
Leaked units & 375 & 0 & 0 & 0 & 0 & 0 & 0 \\
Empty responses & 1 & 2 & 2 & 2 & 2 & 44 & 44 \\
Precision, per query (non-empty) & 18.7\% & 16.4\% & 19.6\% & 23.7\% & 28.5\% & 36.2\% & 36.6\% \\
Precision, pooled & 18.0\% & 16.5\% & 19.8\% & 23.9\% & 28.7\% & 23.1\% & 23.7\% \\
Permitted reached (of 134) & 131 & 131 & 118 & 116 & 102 & 102 & 105 \\
Permitted-file recall & 0.97 & 0.97 & 0.84 & 0.79 & 0.66 & 0.71 & 0.74 \\
\bottomrule
\end{tabular}
\end{table}

At equal output size the ranked filter is comparable to intent routing. The top-3 filter delivers 2.9 units per query, as Context Kubernetes delivers 2.8, and reaches 116 of the 134 answerable queries against 105, at the same pooled precision (23.9\% against 23.7\%). Context Kubernetes leads on per-query precision (36.6\% against 23.7\%), but that average leaves out its 44 empty responses. The filter's ranking contains ties, so we also broke them at random: over 200 random tie-breaks the top-3 filter reaches 115 to 125 queries with pooled precision of 24.0--26.4\% (path order, used in the table, sits at or just below the bottom of both ranges), so the comparison does not depend on how ties fall. On the 47 queries the margin is within that noise: the top-3 filter reaches 29 of 35 (27 to 32 over random tie-breaks) against 28. On this corpus, with a keyword classifier that is 63\% accurate on these queries, intent routing buys little that a ranked access-control filter does not. Whether it does on a real corpus with a better classifier is open (Section~\ref{sec:llm_tension}).

\textbf{Where leakage comes from.} A permission engine that is correct over the labels it receives gives no protection when those labels are assigned by connector configuration rather than derived from the source of record. In our prototype one omitted field sent HR policy to sales agents in most queries while every permission check passed, and the engine-level test (C2) could not have noticed. Labels should come from source metadata, and leakage should be tested end to end, through the connectors. The unscoped leak counts and the queries-with-a-leak row come from \nolinkurl{benchmarks/exp_v2_connector_scoping.py}, the share of queries with a leak under ungoverned retrieval from \nolinkurl{benchmarks/exp_v2_baselines_supplement.py}, and the remaining entries of Tables~\ref{tab:gov_vs_ungov} and~\ref{tab:fourway} from \nolinkurl{benchmarks/exp_v2_matched_baselines.py}. The connector breaks ranking ties by file path, so these numbers do not depend on filesystem order. Because the connector also matches query words against absolute file paths, the name of the temporary checkout can occasionally add a match; in one of our runs this added two units on the 47 queries.

\textbf{V2: Content-level freshness scenarios.} A separate synthetic repository contains an outdated price list, the profile of a churned client marked ``do not contact'', and two status reports for the same project, one ``on track'' and a later one ``at risk''. Four queries are run through the governed pipeline. The status query returns both reports together, a genuine content-level failure: the agent receives contradictory information with nothing to tell it which report is current. The churned client's profile is returned for a client-list query and for a contact query; since the profile itself says ``do not contact'', returning it is arguably correct. (The price-list query returned no units, so it tests nothing.) Source-level freshness cannot catch such failures: every source here is up to date, and the problem is in the content. The prototype's reconciliation loop tracks source freshness (C3) and has no mechanism yet for deleted, superseded or contradictory documents, so we report no governed comparison. Content-level freshness, conflict detection in particular, is the next component to build (Section~\ref{sec:future}).

\textbf{V3: Three-tier vs.\ flat permissions.} Five attack scenarios are submitted, one action each, to three permission profiles (Table~\ref{tab:attack}): no governance (the agent inherits every user permission, all autonomous), basic RBAC (the role's permissions, with the agent excluded from writes), and Context Kubernetes (per-agent tiers). An action counts as gated if it is denied or requires approval. The outcomes follow from how each profile is configured; the experiment shows what the three-tier model can express, not how often a deployment would block a real attack.

\begin{table}[H]
\centering
\caption{Attack scenarios: three permission models compared}
\label{tab:attack}
\small
\begin{tabular}{@{} p{6.5cm} c c c @{}}
\toprule
\textbf{Attack scenario} & \textbf{No gov.} & \textbf{RBAC} & \textbf{CK8s} \\
\midrule
Send email with confidential pricing & Allowed & Allowed & \textbf{T3 approval} \\
Access HR salary data from sales session & Allowed & Blocked & Blocked \\
Sign contract autonomously & Allowed & Blocked & Blocked \\
Access finance records from sales role & Allowed & Blocked & Blocked \\
Modify client records autonomously & Allowed & Blocked & Blocked \\
\midrule
\textbf{Actions gated} & \textbf{0/5} & \textbf{4/5} & \textbf{5/5} \\
\bottomrule
\end{tabular}

\par\vspace{0.5em}
\noindent\begin{minipage}{\linewidth}
\footnotesize\textit{Note:} ``T3 approval'' means the action is held for out-of-band strong approval rather than denied. The Context Kubernetes profile requires it for every \texttt{send\_email} by the agent; the request carries no message content, so the model does not distinguish an email containing pricing from one that does not.
\end{minipage}
\end{table}

The row that separates the models is the first. RBAC asks whether the role may send email (yes). The three-tier model asks whether the \emph{agent} may send email \emph{autonomously}, and here it may not: the action waits for the user's approval over a separate channel. Per-agent tiers let a deployment gate an action the user's role permits, which role-based permissions cannot express: RBAC can deny the action, but not allow it subject to out-of-band approval. The outcomes are computed by \nolinkurl{benchmarks/exp_c_attack_scenarios.py}.

\subsection{Gap Analysis}\label{sec:gap_analysis}

We assess six approaches against the seven requirements (Section~\ref{sec:requirements}), grounded in NIST AI RMF categories. For Context Kubernetes, ``Impl.'' means implemented in the prototype and exercised by the experiments above, on synthetic data; ``Partial'' means only part of the requirement is implemented.

\begin{table}[H]
\centering
\caption{Governance gap analysis (NIST AI RMF--grounded requirements)}
\label{tab:eval}
\small
\begin{tabularx}{\textwidth}{X c c c c c c}
\toprule
\textbf{Requirement} & \rotatebox{70}{\textbf{CK8s (impl.)}} & \rotatebox{70}{\textbf{MS Copilot}} & \rotatebox{70}{\textbf{SF Agentforce}} & \rotatebox{70}{\textbf{AWS Bedrock}} & \rotatebox{70}{\textbf{Google Vertex}} & \rotatebox{70}{\textbf{DIY (OSS)}} \\
\midrule
R1: Vendor neutral & Design & Gap & Gap & Partial & Partial & Addr. \\
R2: Declarative mgmt & Impl. & Gap & Gap & Gap & Gap & Gap \\
R3: Agent perm sep. & Partial & Partial & Partial & Partial & Partial & Gap \\
R4: Context freshness & Partial & Partial & Partial & Gap & Gap & Gap \\
R5: Intent-based access & Partial & Partial & Partial & Gap & Partial & Gap \\
R6: Full auditability & Impl. & Addr. & Addr. & Addr. & Partial & Gap \\
R7: Org.\ intelligence & Design & Gap & Partial & Gap & Gap & Gap \\
\bottomrule
\end{tabularx}
\vspace{0.3em}
\footnotesize\emph{CK8s: ``Impl.'' = implemented and exercised in the prototype; ``Partial'' = R3 separates agent from user permissions for actions but not for reads, R4 detects source-level staleness but neither applies it on delivery nor covers deleted or contradictory content (V2), R5 routes by intent but read authorization compares role labels only (Section~\ref{sec:limitations}); ``Design'' = specified, not shown (R1: the prototype has Git and PostgreSQL connectors and has not been run against several vendors' stacks; R7: the organizational intelligence module is designed but not implemented). R2 means a declarative manifest for the knowledge layer, not infrastructure-as-code for deploying agents. Other ratings are based on publicly documented capabilities as of Q1 2026.}
\end{table}

\section{Related Work}\label{sec:related}

\textbf{Container orchestration.} Kubernetes \citep{burns2016borg, verma2015borg}, Docker Swarm, and Mesos \citep{hindman2011mesos} established the declarative orchestration paradigm for compute. Kubernetes Operators \citep{dobies2020operators} introduced domain-specific controllers. Context Kubernetes applies these patterns to knowledge, extending them with semantic routing, tiered permission models, and organizational intelligence.

\textbf{Distributed systems foundations.} Lamport's work on event ordering \citep{lamport1978clocks} and coordination services \citep{hunt2010zookeeper} inform the registry and reconciliation design. The desired-state model draws on the Borg control plane \citep{verma2015borg}.

\textbf{Role-based access control.} Sandhu et al. \citep{sandhu1996rbac} formalized RBAC for human identities. The subset idea is not new: role-based delegation models already bound a delegate by its delegator's permissions. What we add is a strict-subset requirement for autonomous agents (Design Invariant~\ref{inv:subset}) combined with out-of-band approval tiers on the operations they share.

\textbf{Enterprise agentic platforms.} Microsoft Copilot Studio \citep{microsoft2026copilot}, Salesforce Agentforce \citep{salesforce2026atlas, salesforce2026architecture}, AWS Bedrock Agents \citep{aws2026bedrock}, Google Vertex AI \citep{google2026a2a}, and open-source frameworks \citep{langgraph2025, crewai2025, autogen2025} provide agent orchestration but not vendor-neutral context governance. Abou Ali et al.~\citep{abouali2025agenticai} survey agentic AI architectures. Adimulam et al.~\citep{adimulam2026orchestration} analyze multi-agent orchestration protocols. Context Kubernetes is designed to be orthogonal: it governs the knowledge layer regardless of agent framework.

\textbf{Agent governance and trust.} NeMo Guardrails \citep{nvidia2025nemo} provides I/O filtering. Langfuse \citep{langfuse2025} and LangSmith \citep{langsmith2025} provide observability. Raza et al.~\citep{raza2025trism} survey trust and security management for multi-agent systems. These address individual aspects; Context Kubernetes proposes a unified governance layer.

\textbf{AI reliability certification.} Mouzouni \citep{mouzouni2026trustgate} introduces black-box reliability certification via self-consistency sampling and conformal prediction \citep{vovk2005algorithmic, angelopoulos2023conformal}. Context Kubernetes integrates this for operator deployment gating.

\textbf{Knowledge management.} Nonaka and Takeuchi \citep{nonaka1995knowledge} established organizational knowledge creation theory. Alavi and Leidner \citep{alavi2001knowledge} surveyed KM systems. Context Kubernetes addresses the infrastructure for making organizational knowledge \emph{agent-consumable}---a challenge prior KM systems, designed for human consumption, did not address.

\textbf{Data governance and data mesh.} Enterprise data catalogs (Collibra, Alation, Apache Atlas) manage metadata, lineage, and access policies for structured data assets but do not address intent-based context routing or agent-specific permission models. Dehghani's data mesh paradigm \citep{dehghani2022datamesh}, grounded in domain-driven design \citep{evans2003ddd}, shares structural principles with Context Kubernetes: domain-oriented ownership, declarative governance, and self-serve infrastructure. Context Kubernetes extends these principles to agent-consumable knowledge with semantic routing and organizational intelligence accumulation. Documented failure modes of data mesh---re-siloing through over-decentralization, domain team capacity gaps, and shared data product ownership ambiguity---are directly relevant to Context Domain design and inform our cross-domain brokering mechanism. Federated knowledge graph techniques \citep{chen2021fede} demonstrate privacy-preserving pattern sharing across distributed knowledge stores, informing the design of cross-domain intelligence sharing in Context Operators.

\textbf{Context engineering.} The emergence of context engineering as a discipline \citep{cognizant2026context, infoworld2026context, vishnyakova2026context} reflects growing interest in systematic approaches to organizing enterprise knowledge for AI consumption. MCP \citep{anthropic2025mcp} and A2A \citep{google2026a2a} are transport protocols. Context Kubernetes operates at the governance layer above them.

\textbf{AI risk management.} The NIST AI Risk Management Framework \citep{nist2023airm} provides a structured approach to AI governance. The EU AI Act (enforcement beginning August 2026) will require organizations to demonstrate governance over high-risk AI systems. Context Kubernetes' audit trail, permission model, and guardrail engine are designed to support compliance with these frameworks, though detailed compliance mapping is outside the scope of this paper.

\section{Discussion}\label{sec:discussion}

\subsection{Limitations}\label{sec:limitations}

The prototype and its evaluation have the following limitations.

\textbf{Small synthetic corpus.} Every experiment uses eight synthetic documents. The 200-query set consists of the 47 original queries and 153 templated paraphrases over the same files, so the queries are not independent, and we report no confidence intervals. The value experiments show mechanisms and failure modes; they do not estimate how a deployment would perform.

\textbf{Routing.} The experiments use keyword-based intent classification (70.2\% domain accuracy on 47 queries, 63\% on 200). With correctly labelled connectors, routing quality bounds recall: misrouted queries return nothing relevant (Tables~\ref{tab:gov_vs_ungov} and~\ref{tab:fourway}).

\textbf{Labels depend on configuration.} Units carry the authorized roles of the connector that fetched them, so a misconfigured connector silently mislabels everything it returns (V1). A production implementation should derive labels from source-of-record metadata.

\textbf{Read authorization is label-only.} When delivering context, the prototype checks the session's role against each unit's labels. The agent permission profile, with its resource patterns and tiers, is consulted for actions only. Design Goal~\ref{goal:safety}(1) asks for more: in V1 the agent's profile lists only client and sales resources, yet 10 of the 147 units the corrected pipeline delivers come from the delivery folder, whose units carry a sales-readable label although the role's registered resources do not include it. Design Invariant~\ref{inv:subset} and the approval tiers are therefore implemented for actions, not for reads.

\textbf{Freshness is detected, not enforced.} The reconciliation loop tracks source freshness (C3), but no production code path writes those states onto the units the router delivers: every unit carries the default ``fresh'' state, so the router's freshness filter never removes anything. Design Goal~\ref{goal:safety}(2) and requirement R4 are therefore specified and partly built, not enforced end to end.

\textbf{Retrieval harness.} The Git connector matches query terms against file contents and absolute file paths, and ranks candidates by a crude count of term hits, with ties broken by file path. Retrieval quality in all pipelines is correspondingly low.

\textbf{No production deployment.} All experiments run against seed data in a local prototype. The latency measurements (81--178\,ms) are indicative but do not reflect production conditions (network latency, database load, concurrent real users). A production deployment is the natural next step.

\textbf{Single-author evaluation.} The gap analysis (Table~\ref{tab:eval}) and the experiment design are by the author. The approval isolation survey (Table~\ref{tab:approval_survey}) cites vendor documentation, but independent replication by practitioners with platform deployment experience would strengthen the findings.

\textbf{Design invariants tested, not formally verified.} The 92 tests check invariant enforcement empirically (including adversarial attempts). We use ``design invariant'' and ``design goal'' to distinguish these from formally verified properties. The repository contains a TLA+ specification of the reconciliation loop and an Alloy model of the permission model. A TLC run completed without violation, but in that specification the two safety flags are never set by any action, so the safety invariants hold trivially; only the type invariant was checked. A revised specification that records the true age of served content is in the repository and has not yet been run through TLC. Worked by hand, and with an explicit-state search of the same model (\nolinkurl{formal/tla_bfs.py}), it admits a violation: after seven clock ticks a request is served from a source whose cached freshness lags the clock, so content one tick past its expiry is delivered. If TLC confirms this trace, the reconciliation loop needs to recompute freshness at serving time. The liveness properties have not been model-checked, and the Alloy model has not been run.

\textbf{Organizational intelligence not implemented.} Requirement R7 (Context Operators with pattern extraction, learning loops) is designed but not implemented in the prototype. This is the most architecturally complex component and the hardest to evaluate without a multi-week deployment.

\subsection{The LLM-in-the-Loop Tension}\label{sec:llm_tension}

The Context Router is designed to use LLM-assisted intent parsing for semantic understanding of agent requests. This creates a tension: \emph{a system designed to provide governance guarantees depends on a non-deterministic, probabilistic component for routing decisions}. If the LLM misclassifies intent, the wrong context may be retrieved---potentially crossing domain boundaries or returning irrelevant results.

Nor is an LLM classifier a measured improvement yet. On our 200 queries, replacing the keyword classifier with gpt-4o-mini raised domain accuracy from 63\% to 75\% and added about 1.4\,s of latency per query (\nolinkurl{benchmarks/results_track1.json}). Its effect on retrieval cannot be read from that run: it used the unscoped connectors of the first evaluation, under which routing a human-resources query correctly to the HR domain removes everything a sales agent may see, so a better classifier can lower retrieval scores for reasons unrelated to its quality. Measuring the LLM classifier with correctly labelled connectors is future work.

We identify three architectural mitigations, each with trade-offs:

\begin{enumerate}[leftmargin=2em]
  \item \textbf{Deterministic fallback.} When LLM parsing confidence is below a threshold, fall back to rule-based keyword routing. Trade-off: reduced routing quality for ambiguous queries.
  \item \textbf{Intent classification caching.} Cache LLM classifications for repeated query patterns. Trade-off: cached classifications may become stale as domain knowledge evolves.
  \item \textbf{Small, specialized models.} Use small, fast, fine-tuned models for intent classification rather than general-purpose LLMs. Trade-off: requires training data and maintenance.
\end{enumerate}

None of these mitigations fully resolves the tension. We consider this an open problem: how to build governance-grade infrastructure that incorporates probabilistic AI components without undermining its own guarantees. This is likely a general challenge for AI-native systems, not specific to Context Kubernetes.

\subsection{Organizational Intelligence Governance}\label{sec:intelligence_governance}

Context Operators accumulate cross-organizational patterns from aggregate agent activity. This raises governance questions that go beyond traditional data governance:

\begin{itemize}[leftmargin=2em]
  \item \textbf{Emergent sensitivity.} Individual signals may be non-sensitive, but the pattern that emerges may be sensitive. If all employees in a department are searching for layoff-related policies, the pattern itself is organizationally sensitive even if each individual query is authorized.
  \item \textbf{Memory auditability.} Who audits what the operator has ``learned''? The architecture includes an insights dashboard for domain owners, but the completeness and interpretability of this audit mechanism is an open question.
  \item \textbf{Legal discoverability.} Can an operator's accumulated knowledge be subpoenaed? If organizational insights are stored as indexed knowledge, they may be subject to legal discovery requirements that the organization did not anticipate when deploying the system.
  \item \textbf{Feedback loops.} If operators influence agent behavior (through proactive insights), and agent behavior is what operators learn from, self-reinforcing feedback loops may amplify biases or errors. The minimum signal threshold ($\theta \geq 3$) and decay mechanisms are designed to mitigate this, but their effectiveness is an empirical question.
\end{itemize}

These questions map to specific legal frameworks that any deployment must address:

\begin{itemize}[leftmargin=2em, itemsep=0.2em]
  \item \textbf{GDPR Article 22} prohibits decisions based solely on automated profiling that significantly affect individuals; where organizational insights influence individual agent behavior, the system must ensure either genuine non-attributability or meaningful human review and explicit consent under Article~22(2).
  \item \textbf{eDiscovery obligations} (US FRCP Rules 26, 34): organizational insights stored as indexed knowledge constitute electronically stored information subject to litigation hold and production obligations; retention and purge policies must account for legal discoverability from deployment day one.
  \item \textbf{NLRA Section~7 and NLRB guidance} (GC Memo 23-02): aggregate pattern detection from employee agent activity implicates protections against surveillance that chills protected concerted activity; a system that detects ``all employees in department X are searching for layoff-related policies'' must not enable retaliation against employees exercising labor rights.
  \item \textbf{EU AI Act} (Regulation 2024/1689, Annex III \S4(b)): a context orchestration system whose organizational intelligence module monitors or evaluates employee behavior through aggregate pattern analysis is likely classified as \emph{high-risk}, triggering comprehensive compliance obligations including risk management (Art.~9), transparency (Art.~13), human oversight (Art.~14), and a right to explanation.
\end{itemize}

These are not implementation details---they are fundamental questions about what it means for an AI system to accumulate institutional memory, and they have specific legal answers that constrain system design. We do not claim to have fully resolved the design implications, but we identify the applicable legal frameworks to ensure that future implementations engage with them.

\subsection{Security Considerations}\label{sec:threats}

A context orchestration system is a high-value target: it mediates access to all organizational knowledge. Key threats include:

\begin{itemize}[leftmargin=2em]
  \item \textbf{Prompt injection via context.} Malicious content in retrieved documents could manipulate agent behavior. In coding agents, \citet{mouzouni2026exploitation} found that a single sentence framing the task as a search for hidden things raised the rate at which agents reached for a planted backdoor to over a third of trials on one model, despite an explicit instruction to respect all rules; prompt-level rules alone are a weak control. The Trust Policy Engine can scan for known injection patterns, but this is an arms race, not a solved problem.
  \item \textbf{Connector compromise.} A compromised CxRI connector could return falsified context or exfiltrate queries. Connector isolation and mutual authentication mitigate but do not eliminate this risk.
  \item \textbf{Cross-domain inference.} An agent may infer restricted information from the \emph{absence} or \emph{shape} of brokered responses, even if the content is properly filtered. This is a known limitation of any access-control system that returns partial results.
  \item \textbf{Operator knowledge poisoning.} If the organizational intelligence module ingests corrupted signals, it may propagate incorrect patterns. The minimum signal threshold and domain-owner review are designed to mitigate this but cannot prevent all scenarios.
\end{itemize}

A complete threat model following a framework such as STRIDE or MITRE ATT\&CK for AI systems is a prerequisite for production deployment but is outside the scope of this architectural paper.

\paragraph{Deployment preconditions for design invariants.} Design Invariants~\ref{inv:subset} and~\ref{inv:isolation} and Design Goals~\ref{goal:safety} and~\ref{goal:liveness} hold under the following deployment assumptions, which must be treated as prerequisites rather than optional hardening measures:
\begin{enumerate}[leftmargin=2em, itemsep=0.1em]
  \item Mutual TLS between all middleware services and between the middleware and local agents.
  \item CxRI connectors authenticate to source systems with credentials stored in a secrets vault; connectors are not writable by agents.
  \item The middleware runtime (Permission Engine, Trust Policy Engine, Context Registry) is not compromised---standard infrastructure security applies.
  \item LLM providers operate under data processing agreements that prohibit training on customer prompts.
  \item The out-of-band approval channel (Tier~3) uses a separate authentication factor delivered via a channel that is architecturally inaccessible to the agent's process (e.g., a 2FA app on a personal device, not a browser session the agent controls).
\end{enumerate}
If any of these preconditions is violated, the corresponding invariant or design goal cannot be assumed to hold.

\subsection{The Emergence of Context Engineering}\label{sec:context_engineering}

DevOps emerged because deploying containers at scale required a new operational discipline. \textbf{Context Engineering} is the operational discipline for knowledge infrastructure: designing context architectures, defining permission models, managing context migration, operating reconciliation loops, and governing organizational intelligence.

This role does not exist today in most organizations. If agent deployments grow as container deployments did, it may become a discipline of its own.

\subsection{Future Work}\label{sec:future}

Four directions would most strengthen the contribution. First, \textbf{labels from the source of record}, with read authorization that consults the agent profile, closing the gaps that V1 and the read-authorization limitation expose. Second, \textbf{content-level freshness}: detecting deleted, superseded and contradictory documents, which the reconciliation loop does not cover (V2). Third, \textbf{evaluation at realistic scale}: a real corpus, independent relevance labels, and a pilot deployment, which would show whether intent routing earns its place over a ranked access-control filter. Fourth, \textbf{formal verification}: model-checking the revised TLA+ specification and running the Alloy model.

\section{Conclusion}\label{sec:conclusion}

\noindent We have introduced Context Kubernetes, an architecture for enterprise knowledge orchestration in agentic AI systems, implemented it as an open-source prototype, and evaluated it on a small synthetic corpus.

The result that holds most firmly is a cautionary one. In our first evaluation, a single unscoped connector setting made the governed pipeline leak HR and finance content into most queries while every permission check passed. A permission engine is only as good as the labels it is given: labels should come from the source of record, not from the connector that fetches a unit. With correct labels, the governed pipeline delivers nothing outside the agent's role, but so does a plain access-control filter, and on our corpus a filter that keeps its top three results reaches more of the queries the agent may answer at the same pooled precision. The case for intent routing has yet to be made on realistic data.

The permission model shows what it can express rather than what it prevents: per-agent approval tiers can hold an action that the user's role permits, such as an agent sending email, for approval over a channel the agent cannot read or write, and no surveyed enterprise platform enforces that isolation. The reconciliation loop detects source-level staleness, but the prototype does not yet apply those states to the units it delivers, and content-level freshness, which V2 illustrates, does not yet exist.

Access control for agents is only as good as its labels. Our first evaluation passed every permission test and still sent HR policy to a sales agent; leakage has to be tested end to end, through the connectors that assign the labels.

\section*{Changes from the April 2026 versions}

This version (v4) corrects the evaluation in the three April 2026 versions of this paper. Versions~1 and~2 (13 and 14 April) differ only in minor details; version~3 (16 April) replaced V1 with a four-way comparison on 200 queries and added the TLA+ result. In all three, every connector in V1 was left without a path scope, so the governed pipeline leaked HR and finance content into most queries.
(1)~Versions~1 and~2 reported a 21.6\% leak rate and a 14-point noise reduction as improvements; the noise reduction came from empty responses. Version~3 attributed the governed pipeline's 307 leaks to cross-domain brokering and credited intent routing with a 19-point noise reduction; both came from the unscoped connectors and from empty responses. Its LLM-routing figures were measured on the same configuration.
(2)~All three versions stated in C2 that the leakage bug had been fixed before release; the fix added connector-level labels but not path scoping. Their ``zero unauthorized deliveries'' referred to seven engine-level tests (C2).
(3)~Version~3's attack column in the four-way table was written by hand in the benchmark script, and its claim that RBAC cannot tell email with pricing from email without it was wrong: the three-tier model gates all email. V3 is now reported from \nolinkurl{benchmarks/exp_c_attack_scenarios.py}.
(4)~Version~3 stated that TLA+ model checking verified the safety properties over 4.6 million states; in that specification the safety flags are never set, so the invariants hold trivially.
(5)~V2 is reported as a description of failure scenarios, because its governed arm was a hand-written filter. The code size is 3{,}711 lines (versions~1 and~2 said about 7{,}000), the seed corpus has eight files (not twelve), the connector now breaks ranking ties by file path, and the read-authorization and freshness-enforcement limitations are reported. A detailed list is in \texttt{CHANGELOG-v4.md} in the repository.

\section*{Acknowledgments}

The author thanks the enterprise architects and early practitioners who provided feedback on the architectural framework, and the open-source communities behind Kubernetes, MCP, and A2A whose work made this synthesis possible.

\bibliographystyle{plainnat}

\end{document}